\documentclass[sigconf]{acmart}

\usepackage[inkscapelatex=false]{svg}
\usepackage{subcaption}
\usepackage{wrapfig}
\newcommand{\R}{{\mathbb R}}
\newcommand{\D}{{\mathcal{D}}}
\newcommand{\LL}{{\mathcal{L}}}
\newcommand{\EX}{{\mathbb E}}
\usepackage{graphicx} 
\usepackage[inkscapelatex=false]{svg}

\AtBeginDocument{%
  }

\copyrightyear{2025}
\acmYear{2025}
\setcopyright{acmlicensed}\acmConference[WSDM '25]{Proceedings of the Eighteenth ACM International Conference on Web Search and Data Mining}{March 10--14, 2025}{Hannover, Germany}
\acmBooktitle{Proceedings of the Eighteenth ACM International Conference on Web Search and Data Mining (WSDM '25), March 10--14, 2025, Hannover, Germany}
\acmDOI{10.1145/3701551.3703494}
\acmISBN{979-8-4007-1329-3/25/03}

\begin{document}

\title{Prospective Multi-Graph Cohesion for Multivariate Time Series Anomaly Detection}


\author{Jiazhen Chen}
\email{j385chen@uwaterloo.ca}
\orcid{0000-0001-9962-2974}
\affiliation{%
  \institution{University of Waterloo}
  \city{Waterloo}
  \state{Ontario}
  \country{Canada}
}

\author{Mingbin Feng}
\email{ben.feng@uwaterloo.ca}
\orcid{0000-0002-9748-6435}
\affiliation{%
  \institution{University of Waterloo}
  \city{Waterloo}
  \state{Ontario}
  \country{Canada}}

\author{Tony S. Wirjanto}
\email{twirjanto@uwaterloo.ca}
\orcid{0000-0003-1324-9131}
\affiliation{%
  \institution{University of Waterloo}
  \city{Waterloo}
  \state{Ontario}
  \country{Canada}}

\renewcommand{\shortauthors}{Jiazhen Chen, Mingbin Feng, and Tony S. Wirjanto}

\begin{abstract}

Anomaly detection in high-dimensional time series data is pivotal for numerous industrial applications. Recent advances in multivariate time series anomaly detection (TSAD) have increasingly leveraged graph structures to model inter-variable relationships, typically employing Graph Neural Networks (GNNs). Despite their promising results, existing methods often rely on a single graph representation, which are insufficient for capturing the complex, diverse relationships inherent in multivariate time series. To address this, we propose the Prospective Multi-Graph Cohesion (PMGC) framework for multivariate TSAD. PMGC exploits spatial correlations by integrating a long-term static graph with a series of short-term instance-wise dynamic graphs, regulated through a graph cohesion loss function. Our theoretical analysis shows that this loss function promotes diversity among dynamic graphs while aligning them with the stable long-term relationships encapsulated by the static graph. Additionally, we introduce a "prospective graphing" strategy to mitigate the limitations of traditional forecasting-based TSAD methods, which often struggle with unpredictable future variations. This strategy allows the model to accurately reflect concurrent inter-series relationships under normal conditions, thereby enhancing anomaly detection efficacy. Empirical evaluations on real-world datasets demonstrate the superior performance of our method compared to existing TSAD techniques.

\end{abstract}




\maketitle

\section{Introduction}

Time series anomaly detection (TSAD) is the process of pinpointing unusual or unexpected observations within time series data that deviate significantly from the expected pattern. This research area has extensive applications in numerous industries, including device monitoring, cyber-security, and fraud detection~\cite{anandakrishnan2018anomaly,ren2019time,cook2019anomaly,li2021fluxev,chen2025semi}. In recent years, the proliferation of IoT sensors has resulted in vast amounts of high-dimensional data, elevating the importance of multivariate TSAD. Analyzing relationships among multiple time series allows for the detection of anomalies that may be invisible in isolated univariate analysis. By leveraging these inter-series dependencies, multivariate TSAD can uncover complex patterns and thus enhance the effectiveness of anomaly detection.

Existing TSAD methods can be categorized into supervised and unsupervised learning techniques. Due to the inherent scarcity of labeled anomalies in real-world datasets, unsupervised learning is often preferred, operating under the assumption that all training examples represent normal behavior. Traditional unsupervised anomaly detection methods, such as OCSVM~\cite{ocsvm}, PCA~\cite{shyu2003novel}, and ARIMA~\cite{box2015time}, have solid mathematical foundations and offer good interpretability. However, their performance may be hampered when confronted with large-scale, high-dimensional data. Recent advancements have seen a shift toward deep learning-based methods, which excel in handling various complex data types. These methods primarily utilize neural network architectures such as Convolutional Neural Networks (CNNs) (e.g.,\cite{mscred,tcnanomaly,chen2025harnessing}), Recurrent Neural Networks (RNNs) (e.g.,\cite{lstmed,omnianomaly,lstm-ndt,chen2021daemon}), or Transformers (e.g.,~\cite{anomalyTransformer,li2023prototype,yang2023dcdetector,fang2024temporal}) to process time series data and detect anomalies through forecasting errors or reconstruction errors. Despite their efficacy in capturing temporal dependencies, these neural networks are not inherently designed to explicitly model the relationships among variables, which is crucial for uncovering anomalies in high-dimensional datasets.

Recently, there has been a growing trend in employing Graph Neural Networks (GNNs) to better characterize inter-series relationships within multivariate time series data. These methods typically construct learnable graphs to represent the inter-series relationships and generate rich representations through GNNs based on the learned graphs. For instance, GDN~\cite{gdn}, GTA~\cite{gta}, and FuSAGNet~\cite{fusagnet} construct an inter-series graph based on global learnable embeddings to describe the long-term relationships among time series. Similarly, MTAD-GAT~\cite{zhao2020multivariate},  GReLeN~\cite{zhang2022grelen} and DuoGAT~\cite{lee2023duogat} utilize short-term time series windows to infer these inter-series relationships. Subsequently, they refine the multivariate time series representations through GNNs and optimize under objectives such as forecasting or reconstruction.

However, current GNN-based methods typically rely on a single type of graph, either static or dynamic, to represent relationships among time series. Static graphs capture long-term relationships but lack the adaptability to respond to dynamic changes, while dynamic graphs are often overly sensitive to short-term noise, potentially misinterpreting regular fluctuations as anomalies. For example, in an industrial context, a static graph might mistakenly flag normal operational changes as anomalies due to its inability to adapt to routine fluctuations. Conversely, dynamic graphs, while responsive to short-term changes, may misinterpret long-term sensor variations as significant anomalies, thus triggering false alarms.

To address the aforementioned issues, we introduce the Prospective Multi-Graph Cohesion (PMGC) framework for multivariate TSAD. Specifically, PMGC integrate a list of learnable dynamic graphs with a static graph to capture both the long-standing relationships and evolving short-term interactions among time series. To seamlessly integrate these two types of graphs, we introduce a graph cohesion loss. Our theoretical analysis shows that this loss function encourages the dynamic graphs to remain diverse while converging towards the static graph,  thus ensuring a balance between the cohesiveness and diversity of graph representations. Leveraging these integrated graphs, we employ GNNs to refine the representations of the multivariate time series data, which are optimized under a forecasting objective. Anomalies are subsequently identified based on significant deviations in the forecasting errors.

However, one drawback of traditional forecasting TSAD approaches is their reliance on historical data, which may not adequately capture unpredictable concurrent contextual changes. For example, in manufacturing, production lines might reconfigure workflows during high demand periods, causing short-term changes in process relationships that are normal but unpredictable from historical data alone. To overcome this limitation, we propose a simple but effective strategy called ``Prospective Graphing''. This technique incorporates current time series values into the graph construction process. By incorporating the most recent observations, the graphs are able to more accurately reflect the current inter-series relationships. 

To summarize, our contributions are listed as follows:
\begin{itemize}
\item 
We present a novel framework that integrates instance-specific dynamic graphs with a universal static graph, effectively capturing both long-standing relationships and evolving short-term interactions among time series.

\item We introduce a graph cohesion loss function that ensures diversity among graphs while aligning with the overarching patterns captured by the static graph. Our theoretical analysis demonstrates the effectiveness of this loss function in balancing cohesiveness and diversity in graph representations.

\item We propose a simple yet effective technique called ``Prospective Graphing'', which enhances the traditional forecasting process by taking concurrent inter-series relationships into account.

\item Empirical experiments show that PMGC achieves superior performance on five real-world industrial datasets over a variety of existing TSAD methods.
\end{itemize}



\section{Related Works}

\subsection{Time Series Anomaly Detection}
The mainstream deep learning-based TSAD techniques consist of the forecasting-based and reconstruction-based methods~\cite{garg2021evaluation}. Forecasting-based methods predict future values based on historical data, and anomalies can be identified when substantial deviations between predicted and actual values occur. Existing works for forecasting-based methods generally leverage various neural architectures like RNNs, CNNs, Transformers, or hybrid combinations to learn feature representation of multivariate time series~\cite{lstm-ndt,tcnanomaly,munir2018deepant, zhao2020multivariate,gta,tian2023anomaly,dygraphad}. For example, LSTM-NDT~\cite{lstm-ndt} utilizes Long Short-Term Memory networks (LSTMs) to predict future values and dynamically detects anomalies by analyzing prediction errors with a nonparametric threshold. Similarly, DeepANT~\cite{munir2018deepant} uses CNNs to forecast subsequent time points based on historical data and identifies anomalies from the deviation between predicted and observed values. 

Reconstruction-based methods focus on the accurate reconstruction of time series, under the premise that anomalies will result in high reconstruction errors. Those methods include deep autoencoders (AEs)~\cite{omnianomaly,anomalyTransformer,li2023prototype,chen2021daemon,jie2024disentangled} and Generative Adversarial Networks (GANs)~\cite{li2019mad,tanogan,chen2023adversarial,fang2024temporal} with varying aforementioned sequence encoder architectures to learn the characteristics within time series. For instance, 
MAD-GAN~\cite{li2019mad} integrates a GAN with an LSTM-based reconstruction model, using outputs from both the discriminator and the reconstruction model to determine anomaly scores. AnomalyTransformer~\cite{anomalyTransformer} introduces a dual-branch attention mechanism to enforce long-term dependencies between time points and computes anomaly scores from a combination of association discrepancies and reconstruction loss.

Despite the powerful capabilities of those base neural architectures to capture temporal dependencies, these models are not designed to explicitly model the inter-series relationship. For instance, LSTMs and Transformers typically project input features into dense vectors, which may not explicitly capture the pairwise relationships between different series. Temporal Neural Networks (TCNs)~\cite{tcn}, known for handling temporal sequences, primarily aggregate information through pooling operations, which might not sufficiently address the spatial relationships between multiple series.

\subsection{Modeling Time Series via Graphs}
Thanks to the rapid development of GNNs, deep learning-based methods are capable of modeling graph-structured data. This progress has been leveraged in time series anomaly detection, with an increasing number of studies integrating GNNs into their model design. 
MTAD-GAT~\cite{zhao2020multivariate} utilizes graph attention networks (GATs) with learnable intra- and inter-series graphs to capture spatial-temporal relationships. DuoGAT~\cite{lee2023duogat} extends this by incorporating a directed, weighted, time-oriented graph with a differencing-based GAT layer for temporal dependencies.
GReLeN~\cite{zhang2022grelen} uses a VAE structure to learn a probabilistic graph that captures the dependence relationship among different sensors in a system. STADN~\cite{tian2023anomaly} leverages GATs to model the relationship between sensors based on a dynamic graph constructed using correlation coefficients.

Apart from those approaches that use a short-term time window for graph learning, GDN~\cite{gdn} creates a global static graph using one-hot embeddings and imposes sparsity constraints on it by retaining only top-K entries per node. Similarly, FuSAGNet~\cite{fusagnet} constructs a top-K graph which additionally takes account of external information like categories of sensors, and refines its approach by using sparse auto-encoders to generate node features.
GTA~\cite{gta} further enhances computational efficiency by using Gumbel sampling to create a sparse, differentiable graph, bypassing the need for dot-product multiplication. CST-GL~\cite{zheng2023correlation}, starting with a static learnable global embedding graph, uses a coupled GCN and TCN module to model spatial and temporal dependencies. Other methods include MAD-SGCN~\cite{mad-sgcn}, which adopts self-supervised learning for its graph structure, and MEGA~\cite{wang2022multiscale}, which learns a multi-scale graph based on Discrete Wavelet Transform.
Our method differs from majority of existing approaches by decoupling inter-series relationships into short-term and long-term dependencies through a combination of dynamic and static graphs.

\section{Methodology}
\begin{figure*}[t!]
    \centering
\includegraphics[width=1\linewidth]{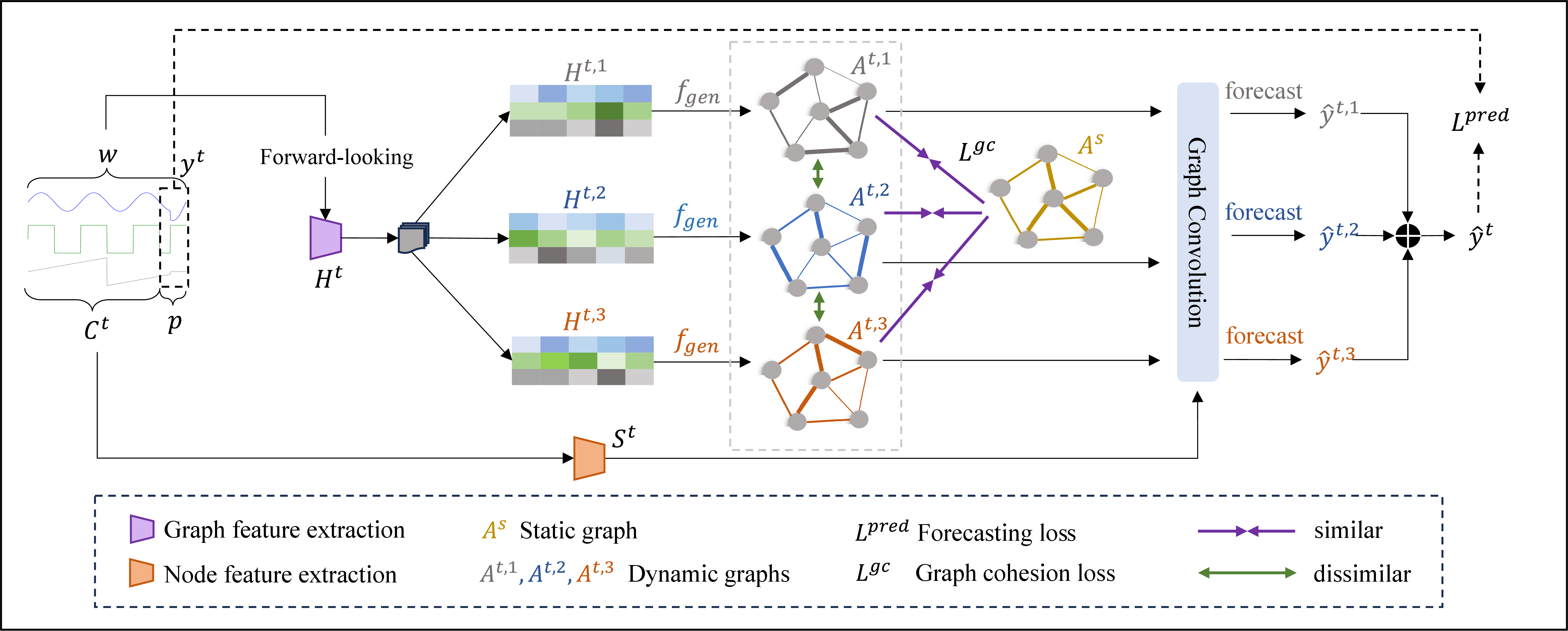} 
    \caption{An illustrative diagram for the Framework of PMGC.}
    \label{fig:framework}
\end{figure*}

\subsection{Problem Formulation}
Following commonly used notations, the $i^{th}$ row and $j^{th}$ column of an arbitrary 2D matrix $A$ are denoted by $A_{i\cdot}$ and $A_{\cdot j}$ respectively. The $(i,j)^{th}$ entry of A is denoted by $A_{ij}$. The $l_2$-norm of a vector is represented by ${\| \cdot \|}_2$, and the Frobenius norm for a matrix is referred as $\|\cdot\|_F$.

The training data is a multivariate time series $X \in \R^{N \times T_\text{train}}$, which consists of $N$ univariate time series collected over $T_\text{train}$ time ticks. We denote $X_{i \cdot}$ as the $i$-th time series and $X_{\cdot t}$ as the time series value at $t$. We adopt the window-based strategy to extract sub-sequences as training or test samples from the multivariate time series by sliding over the multivariate time series along the time dimension with a stride of 1. In other words, the training set $\D$ will be a collection of sub-sequences, where a sample $X^t = \{X_{\cdot i}\}_{i=(t-w+1)}^t \in \D$ is a collection of time ticks over $w$ time steps.

Our goal is to detect anomalies in a test set, a multivariate time series $X \in \R^{N \times T_\text{test}}$ but collected over a separate time span of time ticks. Specifically, we aim to assign an anomaly score to each time tick in the test set, with higher values indicating a greater likelihood of an anomaly. The output of our method is a list of anomaly scores over the $T_{\text{test}}$ time ticks. The anomaly score can be thresholded to a binary value with 0 being normal and 1 being abnormal.

\subsection{Overview}
An overview of the proposed framework, PMGC, is presented in this section. Our framework unfolds in the following steps: \begin{itemize}

    \item Graph Structure Learning (Section~\ref{sec: graph structure learning}): A list of adjacency matrices are derived from the hidden representations of the input time window (Section~\ref{sec: dynamic graph generation}). Subsequently, a graph cohesion loss is imposed on the learned graphs to ensure diversity among them while adhering to certain global properties through a learnable static graph (Section~\ref{sec: graph cohesion}). We provide theoretical analysis to support the effectiveness of the loss function (Section~\ref{sec: theory}).  
    \item Prospective GNN-based Forecasting (Section~\ref{sec: forecasting process}): 
    Future information is incorporated into the graph construction process of dynamic graphs. Subsequently, each dynamic graph is fed into a graph convolution with the encoded context window as node features. The outputs from the graph convolutions are averaged for the time series forecasting. 
    \item Anomaly Score Calculation (Section~\ref{sec: anomaly score generation}): The anomaly score is calculated according to the forecasting error.
\end{itemize}

\subsection{Graph Structure Learning}
\label{sec: graph structure learning}
\subsubsection{Multi-Graph Generation} 
\label{sec: dynamic graph generation}
In this section, we delve into the graph structure learning module, which aims to capture the intricate inter-series relationships in a task-specific manner. Given the evolving nature of real-world time series data, we start by constructing a dynamic graph based on each input time window. Additionally, inspired by the Multi-Head Attention (MHA) mechanism~\cite{transformer}, we embrace a multi-graph approach instead of just generating one graph, under the premise that multiple graphs can be attuned to distinctive aspects of time series data like shapes, frequencies, and granularities. An additional advantage of employing multiple graphs stems from an enhancement in model robustness. Specifically, each graph is learned to capture specific patterns and relationships within the data. This diversification ensures that for an anomaly to go undetected, it would have to simultaneously fit the distinct pattern of all graphs.


Given a time window $X^t$, we initiate the process by extracting its feature representation $H^t$ via a two-layer fully connected neural network. Similar to the MHA mechanism, $H^t$ is then partitioned into $k$ distinct hidden representations $\{H^{t,i}\}_{i=1:k}$, where $H^{t,i}\in \R^{N \times d}$. These representations are then processed through a graph generator function $f_{gen}(\cdot): \R^{N\times d} \rightarrow \R^{N\times N}$ to yield $k$ dynamic graphs $\{A^{t,i}\}_{i=1:k}$. 
There are various ways to perform graph structure learning, such as those based on attention mechanisms~\cite{zhao2020multivariate} or gumbel-softmax techniques~\cite{gta}. In this work, we use the pairwise cosine similarity to generate edge weights, owning to its computational efficiency. Specifically, the operation of $f_{gen}(H)$ is defined as:
\begin{align}
    A^*_{ij} = &\frac{H_{i\cdot}H_{j\cdot}^T}{\|H_{i\cdot}\|_2\|H_{j\cdot}\|_2} \\
   A = &\text{ReLU}(A^*)
\end{align}
Here, each of the $N$ series is treated as a node, with the edge weights calculated based on the representations of pairwise time series. 


\subsubsection{Graph Cohesion Loss} 
\label{sec: graph cohesion}
One drawback of relying solely on those dynamic graphs is their susceptibility to noise and transient fluctuations. To mitigate this issue, we propose a solution by encouraging dynamic graphs to emulate the long-term dependencies captured by a learnable static graph $A^s$. Here, $A^s$ is a static learnable global adjacency matrix, obtained by $A^s = f_{gen}(\xi)$, where $\xi \in \R^{N\times d}$ contains $N$ learnable vectors uniquely defined for each channel of the multivariate time series.

A simple loss formulation to combine the two types of the graphs involves directly minimizing the distance between each dynamic graph and the static graph. We refer this loss function as the simplified version of graph cohesion loss, which is defined as: 
\begin{align}
        \LL^{gc,simple} = \EX_{X^t\sim \D} \left[\sum_{i=1}^{k} \text{dist}(A^s, A^{t,i}) \right]
    \label{eq. graph_cohesion_v1}
\end{align}
Here, $\text{dist}(\cdot, \cdot)$ measures the distance between two graphs, which could be quantified using metrics such as the Frobenius norm. However, directly minimizing the distance between dynamic and static graphs could result in them becoming identical, inadvertently merging into a single graph.

To prevent this, we introduce a new graph cohesion loss which is
designed to maintain long-term dependencies produced by the static graph while preserving the unique patterns of each dynamic graph. The graph cohesion loss is defined as:
\begin{align}
    \LL^{gc} = \EX_{X^t\sim \D} \left[ \sum_{i=1}^{k} -\log \frac{h(A^s, A^{t,i})}{h(A^s, A^{t,i})+\sum_{j\neq i}{h(A^{t,i}, A^{t,j})}}\right]
    \label{eq. graph_cohesion_v1}
\end{align}
where $h(\cdot, \cdot): (\R^{N \times N}, \R^{N \times N}) \rightarrow \R$ is formulated as:
\begin{align}
    \label{eq. similarity_measure}  
      &h(A^i, A^j) = \exp{(-\text{dist}(A^i, A^j)/\tau)} \\
    & \text{dist}(A^i, A^j) = \|A^i - A^j\|^2_F
\end{align}
Here, Eq.~\ref{eq. similarity_measure} measures the similarity between two adjacency matrices with $\tau$ as a temperature hyper-parameter. Intuitively,
$\LL^{gc}$ propels each dynamic graph towards congruence with the global static adjacency matrix $A^s$, ensuring that a universal structure underpins each graph. Concurrently, it encourages diversity among the dynamic graphs, preventing them from converging into identical matrices and ensuring that each retains its unique representational capacity. 


\subsubsection{Theoretical Analysis}
\label{sec: theory}
In this section, we present the theoretical analysis to show that the simplified graph cohesion loss, $\LL^{gc,simple}$, fails to ensure diversity among dynamic graphs and may cause them to converge into the static graph. In contrast, the improved graph cohesion loss, $\LL^{gc}$, does not have the aforementioned issues.

\begin{proposition}
The optimal solution for $\LL^{gc,simple}$ is achieved only when $A^{t,1} = A^{t,2}=\ldots=A^{t,k}=A^s$ for all $X^t \in \mathcal{D}$.
\label{prop. graph cohesion simple}
\end{proposition}

\textbf{Proof.}
When the condition $A^{t,1} = A^{t,2}=\ldots=A^{t,k}=A^s$ is satisfied, it is trivial to show that $\LL^{gc,simple}=0$.
Now assume there exists at least one $A^{t,j}\neq A^s$. In this case, we have $\| A^{t,j}-A^s\|_F^2 > 0$, resulting in $\LL^{gc,simple} > 0$.
Thus, the optimal solution for $\LL^{gc,simple}$ is achieved only when all dynamic graphs collapse into the static graph $A^s$. \qed

Next, proposition~\ref{prop. constant dynamic and static graph} demonstrates that the situation that all dynamic graphs collapse into the static graph is not an optimal solution to minimize $\LL^{gc}$.

\begin{proposition}
The uniform graph case where $A^{t,1} = A^{t,2}=\ldots=A^{t,k}=A^s$ is not an optimal solution to minimize $\LL^{gc}$.
\label{prop. constant dynamic and static graph}

\end{proposition}

\textbf{Proof.}
Given an $X^t \in \mathcal{D}$ and arbitrary $A^s$, define $l_1(X^t)$ as the loss when $A^{t,1} = A^{t,2}=\ldots=A^{t,k}=A^s$. Without loss of generality, define $l_2(X^t)$ as the loss when $\|A^{t,1}-A^s\|_F =d_t > 0$ while the rest remains the same. Additionally, we assume $\tau=1$ for simplicity. Then it follows:
\begin{align}
l_1(X^t) &= \sum^k_{j=1} \left[-\log \left( \frac{1}{k} \right) \right]  = k \log(k) \\
l_2(X^t) &= \log \left( k \right) - (k-1) \log \left( \frac{1}{k-1 + e^{-d_t^2}} \right)
\end{align}

Subtracting $l_2(X^t)$ from $l_1(X^t)$ yields:
\begin{align}
l_1(X^t) - l_2(X^t) &= k \log(k) - \log(k) + (k-1) \log \left( \frac{1}{k-1 + e^{-d_t^2}} \right) \\
&= (k-1) \log \left( \frac{k}{k-1 + e^{-d_t^2}} \right) = \Delta_{d_t}
\end{align}

Given that $e^{-d_t^2} < 1$, $k>1$ and $\log(\cdot)$ is monotonically increasing, it follows that $l_1(X^t)-l_2(X^t) = \Delta_{d_t}> 0$.

Let $\mathcal{P}$ be the probability measure over the distribution $\mathcal{D}$. let $\LL_1$ be the overall loss when $\|A^{t,j}-A^s\|_F=0$, for all $j$ and $X^t$. Now, assuming the model is trained such that the condition $\| A^{t,1} - A^s\|_F > d$ ($d>0$) holds for a subset of $\mathcal{D}$, by which we denote the set as $\mathcal{B}$. Defining the overall loss for this scenario as $\LL_2$, it follows:
\begin{align}
     \LL_1 - \LL_2 &= \int{l_1(X^t) d\mathcal{P}} - (\int_{\mathcal{B}}{l_2(X^t) d\mathcal{P}} + \int_{\mathcal{B}^c}{l_1(X^t) d\mathcal{P}})  \label{eq:difference between constant case}  \\
     &= \int_{\mathcal{B}} l_1(X^t) d\mathcal{P} - \int_{\mathcal{B}}{l_2(X^t) d\mathcal{P}} \\
     & >  \mathcal{P}(\mathcal{B})\Delta_d \label{eq:L1 L2 bound}
\end{align}
 Since $\Delta_d > 0$ and $\mathcal{P}(\mathcal{B})>0$, it follows that the uniform graph case is not the optimal solution since $\LL_1 > \LL_2$. Furthermore, by increasing the number of samples that satisfy $\|A^{t,1}-A^s\|_F \geq d$, $\mathcal{P}(\mathcal{B})$ increases, thereby further minimizing the bound in Eq.~\ref{eq:L1 L2 bound}. This ultimately avoids the uniform graph scenario for all $X^t$.
 \qed

Next, Proposition~\ref{prop. constant dynamic graph} shows that even if the dynamic graphs are identical (but not necessary equal to $A^s$), they still do not minimize $\LL^{gc}$. Thus guarantees the diversity among dynamic graphs.

\begin{proposition}
    The homogeneous dynamic graph case where $A^{t,1} = A^{t,2}=\ldots=A^{t,k}$ does not minimize $\LL^{gc}$.
    \label{prop. constant dynamic graph}
\end{proposition}

\textbf{Proof.}
Building on top of Proposition~\ref{prop. constant dynamic and static graph}, it remains to show that the homogeneous dynamic graph case does not generate better solution in comparison to uniform graph scenario in Proposition~\ref{prop. constant dynamic and static graph}. 

Given an $X^t \in \mathcal{D}$ and an arbitrary $A^s$, assume that $A^{t,1} = A^{t,2} = \ldots = A^{t,k} = A^{t,*}$, where $A^s \neq A^{t,*}$. Define $\|A^s - A^{t,*} \|_F = c_t$. The loss under this setting, $l_3(X^t)$, is calculated as:
\begin{align}
l_3(X^t) &= \sum_{j=1}^k -\log{\left( \frac{e^{-{c_t}^2}}{e^{-{c_t}^2}+(k-1)} \right)} \\
&= \sum_{j=1}^k -\log{\left( \frac{1}{1+\frac{k-1}{e^{-{c_t}^2}}} \right)} = \Delta_{c_t}
\end{align}
Since $\frac{k-1}{e^{-{c_t}^2}} > k-1$ as $e^{-{c_t}^2} < 1$, we have:
\begin{align}
l_3(X^t) > \sum_{j=1}^k -\log{\left( \frac{1}{k} \right)} = l_1(X^t)
\end{align}

Now, define $\mathcal{L}_1$ as the overall loss where $\|A^s - A^{t,*}\| = 0$ for all $X^t \in \mathcal{D}$, and $\mathcal{L}_3$ as the loss where $\|A^s - A^{t,*}\| > c$ with $c > 0$ for some subset of $\mathcal{D}$ (denote this set as $\mathcal{C}$). Similarly to Eq.~\ref{eq:difference between constant case}, it can be shown that:
\begin{align}
\mathcal{L}_3 - \mathcal{L}_1 > \mathcal{P}(\mathcal{C}) \Delta_c
\end{align}
Since $\Delta_c > 0$ and $\mathcal{P}(\mathcal{C}) > 0$, it follows that $\mathcal{L}_3 > \mathcal{L}_1$. This shows that the homogeneous dynamic graph case does not yield a better solution compared to the uniform graph case. Since in Proposition~\ref{prop. constant dynamic and static graph}, it was shown that the uniform graph case is not optimal, it follows that the configuration where $A^{t,1} = A^{t,2} = \ldots = A^{t,k}$ is also not an optimal solution to minimize $\mathcal{L}^{gc}$. \qed

\subsection{Prospective GNN-based Forecasting}
\label{sec: forecasting process}
In this study, we follow the routine of the forecasting-based TSAD approach, which identifies anomalies by measuring deviations between predicted and actual values. Traditional forecasting-based TSAD methods often rely on historical patterns, which may fail to account for concurrent contextual changes. For instance, in manufacturing, different batches of products might require unique processing settings, and during high demand periods, production lines may be temporarily reconfigured. These changes introduce dynamic inter-process relationships that are normal but unpredictable using historical data alone.

To address this, we introduce a technique termed as ``prospective graphing'', which incorporates current time steps into the graph modeling instead of solely using them as labels. By using up-to-date information to construct dynamic graphs, the model adjusts in real-time, accurately reflecting current inter-series relationships. Unlike reconstruction-based TSAD methods, which suffers from issues like identity mapping, our approach solely incorporates current time into the graph construction process. The predictions are then based on how these graphs aggregate historical observations, considering current inter-series relationships. Additionally, these dynamic graphs are further regularized by $\LL^{gc}$, mitigating the identity mapping issue and enhancing model robustness.

Next, GNNs are used as the feature encoder to aggregate contextual information for each time series based on these prospective dynamic graphs. The extraction of node features is done by applying linear layers to raw time series data. As highlighted in~\cite{DLinear}, this simplified approach has competitive performance on time series forecasting tasks in comparison to other advanced neural architectures (e.g., Transformers and RNNs), offering an optimized balance between performance and computational efficiency.

Concretely, the input sample $X^t$ is split into a context window $C^t=\{X_{\cdot i}\}_{i=(t-w+1)}^{(t-p)}$ and a prediction window $y^t=\{X_{\cdot i}\}_{i=(t-p+1)}^t$. The context window which is encoded by the linear layers, together with the dynamic graphs $\{A^{t,i}\}_{i=1}^k$, is fed into GNNs in turn to generate a list of hidden features which are decoded into the $p$-step future sequence, $\hat{y}^t$. The operations unfold as follows:
\begin{align}
   & S^{t} = f_{e}{(C^t)} \\
   & E^{t,i} = f_{gnn}(S^t, A^{t,i}) \\
   & \hat{y}^{t,i} = f_d(E^{t,i})   \\
   & \hat{y}^t = \frac{1}{k} \sum_{i=1}^k{\hat{y}^{t, i}}
\end{align}
where $f_{e}(\cdot):\R^{(w-p)} \rightarrow \R^{d}$ is a linear layer that encodes the context vector into its hidden representations, and $f_{d}(\cdot):\R^{d} \rightarrow \R^{p}$ is a linear layer that decode the output of GNNs into the final predictions. We adopt MIXHOP~\cite{wu2020connecting} as the graph convolution layer for $f_{gnn}(Z, A)$. The operation of each layer is defined as:
\begin{align}
    Z = \beta Z + (1-\beta)\tilde{A}ZW 
\end{align}
where $\beta$ controls the ratio that retains the original states, $\tilde{A}$ is the symmetric normalized Laplacian matrix of $A$, and $W$ is a learnable weight matrix. The predicted value $\hat{y}^t$ is obtained from an averaging of the predictions $\hat{y}^{t,i}$ derived from different dynamic graphs. 

The prediction loss is defined as the mean squared error:
\begin{align}
    \LL^{pred} = \EX_{X^t \sim \D}{\lVert y^t - \hat{y}^t \rVert}_2^2
\end{align}
where $y^t$ is the true value within the prediction window for $X^t$.
Thus, the overall training loss is a combination of the prediction loss and the graph cohesion loss:
\begin{align}
    \LL=\LL^{pred}+\lambda \LL^{gc}
\end{align}
where $\lambda$ is a hyper-parameter that controls the influence of the graph cohesion loss on the learned graphs.


\subsection{Anomaly Score Calculation}
\label{sec: anomaly score generation}
The anomaly score is calculated based on the forecasted error. However, since different time series have different behaviors which can result in very different scales of forecasted errors, we adopt the same normalization method as proposed by GDN~\cite{gdn}, which leverages the median and interquartile range (IQR) to standardize the forecasting error of each time series. Specifically, the anomaly score for the $i$-th time series at time $t$, denoted as $a_i^t$, is calculated as:
\begin{align}
    a_i^t=\frac{Err_i^t-\tilde{\mu}_i}{\tilde{\sigma}_i}    
\end{align}
Here, $Err_i^t$ represents the forecasted error for time series $i$ at time $t$, while $\tilde{\mu}_i$ and $\tilde{\sigma}_i$ denotes the median and IQR of the error set for that specific time series. The final anomaly score at $t$ is then obtained by taking the maximum among all sensors. Anomalies are then identified based on a predefined threshold: if the anomaly score at time $t$ exceeds the threshold, the instance is flagged as anomalous.


\section{Experiments}

\subsection{Datasets}
In this study, we adopt five widely-used real-world benchmark datasets for performance evaluation: SWaT~\cite{mathur2016swat}, WADI~\cite{ahmed2017wadi}, HAI~\cite{hai}, MSL, and SMAP~\cite{lstm-ndt}. SWaT and WADI datasets contain multivariate time series from water treatment and distribution systems, respectively, with normal operations for the initial period and cyber attack data for the latter. MSL and SMAP, published by NASA, provide telemetry anomaly data from spacecraft operations. The HAI dataset emulates power generation processes, which is collected from an industrial control system testbed. Following previous studies~\cite{gdn}, we down-sample the data from SWaT, WADI, and HAI to one measurement every 10 seconds using the median value. Additionally, each dataset includes an unlabeled training set and a separate test set with labeled anomalies. 


\begin{table*}[t!]
\small
    \setlength{\abovecaptionskip}{0cm}
    \centering
        \caption{Performance comparison based on point-wise F1 (\%) as F1$^p$, F1-composite (\%) as F1$^c$, VUS-ROC(\%) as ROC and VUS-PR(\%) as PR. The best performances are highlighted in bold, while the second and third best are underlined.}
   \resizebox{\linewidth}{!}{
    \begin{tabular}{ccccccccccccccccccccc}\\ \toprule
               &   \multicolumn{4}{c}{\textbf{SWaT}} & \multicolumn{4}{c}{\textbf{WADI}} & \multicolumn{4}{c}{\textbf{MSL}} & \multicolumn{4}{c}{\textbf{SMAP}} & \multicolumn{4}{c}{\textbf{HAI}}\\ \midrule
               \textbf{Methods} & \textbf{F1}$^{p}$ &  \textbf{F1}$^c$ & \textbf{PR} &  \textbf{ROC}& \textbf{F1}$^{p}$ &  \textbf{F1}$^c$ 
               & \textbf{PR} &  \textbf{ROC}& \textbf{F1}$^{p}$ &  \textbf{F1}$^c$& \textbf{PR} &  \textbf{ROC} & \textbf{F1}$^{p}$ &  \textbf{F1}$^c$ & \textbf{PR} &  \textbf{ROC} & \textbf{F1}$^{p}$ &  \textbf{F1}$^c$ & \textbf{PR} &  \textbf{ROC}  \\ \midrule

MAD-GAN & 73.6 & 44.4 & 44.2 & 70.8 & 15.9 & 36.6 & 10.8 & 47.0 & 20.2 & 28.4 & 15.2 & 55.9 & \underline{24.3} & \underline{28.6} & 14.0 & 50.8 & 5.7 & 7.2 & 3.2 & 50.5 \\MSCRED & 75.9 & 37.1 & 34.9 & 54.5 & 29.6 & 36.4 & 27.2 & 74.9 & \underline{27.9} & 41.7 & 24.1 & \underline{71.3} & 23.0 & 23.6 & 12.2 & 44.7 & 36.9 & 58.0 & 25.2 & 78.4 \\GDN & \underline{77.5} & 50.7 & \underline{52.0} & 80.6 & \underline{49.5} & \underline{64.9} & \underline{43.3} & \underline{80.3} & 23.9 & 41.7 & \underline{22.1} & \underline{68.5} & 23.6 & 31.5 & 14.3 & 52.4 & 42.5 & 58.0 & \underline{35.1} & \underline{88.6} \\GTA & 77.1 & 52.0 & 42.9 & 61.8 & 27.1 & 46.0 & 18.7 & 55.5 & 22.2 & 33.3 & 19.4 & 66.3 & 22.8 & 24.0 & 12.3 & 46.3 & \underline{46.0} & \textbf{72.6}	& \underline{29.4} & 78.2 \\TranAD & 76.9 & 46.4 & 39.1 & 59.1 & 27.3 & 43.1 & 17.5 & 55.0 & 24.6 & \underline{42.3} & 19.8 & 65.0 & 23.5 & 26.3 & 14.5 & 52.7 & \underline{43.5} & \underline{69.9} & 27.1 & 79.0 \\AnomalyTrans & 12.9 & 32.2 & 19.7 & 52.4 & 5.0 & 10.4 & 9.7 & 51.2 & 8.9 & 27.1 & 15.4 & 51.9 & 8.3 & 24.5 & 15.2 & 51.2 & 5.3 & 17.2 & 4.3 & 50.8 \\FuSAGNet & \textbf{81.9} & 55.7 & 58.3 & \underline{83.5} & \underline{48.0} & \underline{59.2} & \underline{39.4} & \underline{76.1} & 
\underline{27.3} & \textbf{64.4} & \underline{25.5} & 64.1 & 22.7 & \underline{43.6} & 15.6 & 50.6 & 35.8 & 48.6 & 24.9 & \underline{86.1} \\GReLeN & 22.9 & 29.7 & 15.0 & 53.9 & 12.5 & 36.6 & 13.0 & 60.6 & 22.2 & 23.9 & 16.8 & 50.7 & \underline{26.3} & 27.3 & \underline{17.2} & 48.3 & 6.8 &	13.9 &	3.9 &	52.8 
\\DCdetector & 21.9 & 22.7 & 14.7 & 55.5 & 11.0 & 15.7 & 9.1 & 55.5 & 19.3 & 25.6 & 14.5 & 54.6 & 24.2 & 26.2 & \underline{16.0} & \underline{57.6} & 3.9 & 4.5 & 2.9 & 54.2
\\CST-GL & 76.8 & \underline{59.1} & \underline{62.2} & \underline{81.7} & 34.1 & 58.2 & 26.6 & 75.7 & 22.8 & 24.4 & 17.9 & 65.7 & 24.7 & 31.3 & 15.7 & \underline{57.0} & 14.6 & 18.4 & 8.0 & 77.6 \\COCA & 67.9 & \underline{58.8} & 48.3 & 75.6 & 20.6 & 34.3 & 15.5 & 56.3 & 22.9 & 36.8 & 20.4 & 66.4 & 24.9 & 26.5 & 14.9 & 51.9 & 32.3 & 57.1 & 23.7 & 77.0  \\ \midrule

PMGC (ours) & \underline{81.2} & \textbf{71.5} & \textbf{65.3} & \textbf{85.6} & \textbf{53.4} & \textbf{66.6} & \textbf{48.1} & \textbf{84.9} & \textbf{33.6} & \underline{58.7} & \textbf{27.8} & \textbf{74.5} & \textbf{30.0} & \textbf{69.4} & \textbf{21.9} & \textbf{65.3} & \textbf{47.7} & \underline{68.7} & \textbf{39.3} & \textbf{89.7} \\
   \bottomrule
    \end{tabular}}
    \label{tab:comparison_result}
\end{table*}


\subsection{Experimental Setup}
\label{sec: experimental_setting}
PMGC is trained with an Adam optimizer~\cite{kingma2014adam} at a learning rate of $10^{-3}$ for 10 epochs. $20\%$ of the training set is selected as a validation set where the best epoch is stored based on the validation loss. The hidden dimension is set to $d=64$. The number of MixHop graph convolution layers is set to 2 with $\beta=0.05$. The prediction window size is set to 5. The weight of the graph cohesion loss $\lambda$ is $10^{-5}$. The number of dynamic graphs $k$ is set to 5. The sequence length of the input sample $w$ is set to 100 and 20 for SMAP and WADI, and 40 for MSL, SWaT, and HAI. All results are averaged over 5 independent runs. 



\subsection{Baseline Models and Evaluation Metrics}
We choose a set of deep learning-based TSAD approaches as the baseline methods, including MAD-GAN~\cite{li2019mad}, MSCRED~\cite{mscred}, GDN~\cite{gdn}, GTA~\cite{gta}, TranAD~\cite{tuli2022tranad}, AnomalyTransformer~\cite{anomalyTransformer}, FuSAGNet~\cite{fusagnet}, GReLeN~\cite{zhang2022grelen}, COCA~\cite{wang2023deep},  DCdetector~\cite{yang2023dcdetector}, and CST-GL~\cite{zheng2023correlation}. 



Acknowledging recent debates about suitable evaluation metrics in this domain~\cite{kim2022towards}~\cite{garg2021evaluation}, we evaluate performance using multiple metrics. These include the widely used point-wise F1 score, which measures precision and recall at each time point, the F1-composite score~\cite{garg2021evaluation}, which provides a balanced measure of time-wise precision and event-wise recall, and VUS-ROC and VUS-PR~\cite{paparrizos2022volume}, which avoid the need to compute thresholds and incorporate the concept of buffer regions designed for range-based contextual anomalies. For the F1 metrics, the threshold for each model is chosen to be the one that generate the best F1, in order to avoid bias towards any specific thresholding method ~\cite{zhang2022grelen}~\cite{gdn}~\cite{fusagnet}. 


\subsection{Comparison Study}
Table~\ref{tab:comparison_result} presents the comparative results of all models. Overall, PMGC achieves the best performance on average across all datasets. Specifically, we outperform the best-performing baseline by 13\% in pointwise F1, 18\% in F1-composite, 8\% in VUS-PR, and 21\% in VUS-ROC. 

For non-graph-based methods such as TranAD, AnomalyTransformer, DCDetector, and COCA, the performance lags is likely attributed to their relatively weaker ability to model inter-series relationships. These methods primarily focus on capturing the temporal pattern within individual time series while the interactions among them is only learned in an implicit way, which limit their effectiveness in capturing the complex dependencies within multivariate time series. For graph-based methods, MSCRED and GReLeN operate directly on dynamic sequential graphs, which is at a coarser scale compared to the raw time series, thus limiting their ability to capture fine-grained temporal dependencies. 

Similar to ours, GDN, FuSAGNet, and CST-GL are graph-based methods utilizing learnable graphs to model relationships between variables and leverage forecasting or reconstruction errors to identify anomalies. However, these methods rely primarily on a single static graph, which may restrict their ability to learn diverse and dynamic evolving relations within multivariate time series. In contrast, PMGC leverages a combination of concurrent dynamic graphs and a static graph, enabling the model to capture both real-time instance-wise short-term relationships and long-term correlations. This comprehensive approach allows anomalies to be detected more effectively if they violate any of these patterns. As a result, PMGC achieves an average improvement of over 10\% across all metrics compared to these competitive baselines, underscoring the effectiveness of the PMGC framework.


\subsection{Ablation Study}

We conduct an ablation study based on the following variants:
\begin{itemize}

\item \textbf{w/o dynamic graph}: Predictions are generated based on the graph convolution outputs of the static graph.
\item \textbf{w/o static graph ($\LL^{gc}$)}: The graph cohesion loss is removed, so only dynamic graphs are utilized in the model.
\item \textbf{w/o prospective graphing}: The construction of dynamic graphs are solely based on historical observations.
\item \textbf{static $\oplus$ dynamic}: $\mathcal{L}^{gc}$ is removed. The outputs of the dynamic graphs and the static graph are averaged to generate the final prediction.
\item \textbf{static $||$ dynamic}: The output embeddings from each dynamic graph's convolution layer are concatenated with those from the static graph. 
\item \textbf{with $\mathcal{L}^{gc, simple}$}: We directly minimize the distance between the learned dynamic graphs and the static graph by replacing $\mathcal{L}^{gc}$ with $\mathcal{L}^{gc, simple}$ (Eq.~\ref{eq. graph_cohesion_v1}).
\end{itemize}

\begin{figure}
    \centering
            \includegraphics[height=5.2cm]{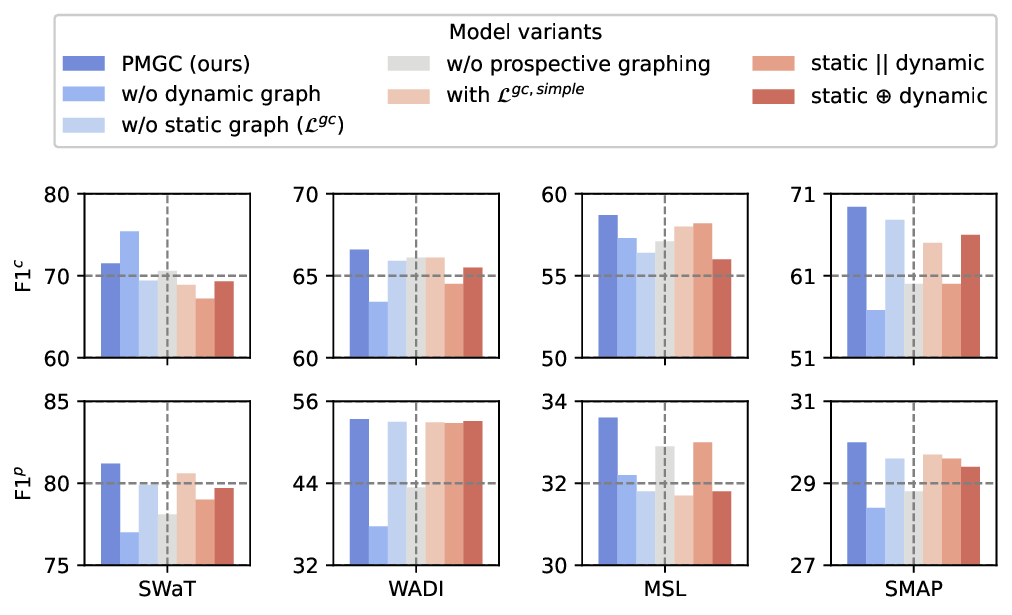}
    \caption{Performance comparison for ablation study in terms of F1$^c$ (\%) (first row) and F1$^p$ (\%) (second row).}
    \label{fig:ablation_result}
\end{figure}

Figure~\ref{fig:ablation_result} presents the performance for possible model variants of MGPC across four datasets. Firstly, removing dynamic graphs on average leads to a noticeable performance drop. Dynamic graphs adapt to changing patterns and interactions within the time series data, which provid a more responsive and accurate modeling. 
Secondly, excluding the static graph (without graph cohesion loss) results in a large performance decline. The reasons are twofold. Firstly, it forces the model to rely solely on short-term information, making it more susceptible to noise. Additionally, it fails to encourage diversity among the dynamic graphs. Using diverse dynamic graphs ensures that anomalies are detected under various graph structures, making it harder for anomalies to have low prediction errors consistently.
Additionally, performance drops significantly without prospective graphing, highlighting the importance of using real-time information in constructing dynamic graphs. This approach ensures that the graphs accurately reflect the current interactions among time series during normal state.

We also compare three variants of ways to combine static and dynamic graphs. Specifically, merely minimizing the distance (by using $\mathcal{L}^{gc, simple}$) between the two graph types can result in overly similar graphs, compromising their ability to preserve the diverse relationships within the time series. Likewise, simply averaging the prediction outputs from each type of graph or concatenating their graph convolution outputs shows a marked performance decline. These methods do not promote diversity among the graphs, thus limiting their ability to capture multifaceted inter-series patterns. Meanwhile, they fail to address the potential drawbacks of the noise in dynamic graphs and the lack of real-time adaptation in static graphs. In comparison, our loss mitigates these issues by ensuring dynamic graphs maintain long-term consistency, thus enhancing the overall quality of the graph representations.

\begin{figure*}[t!]
  \centering
  \begin{subfigure}[b]{1\linewidth}
      \centering
    \includegraphics[height=0.6cm]{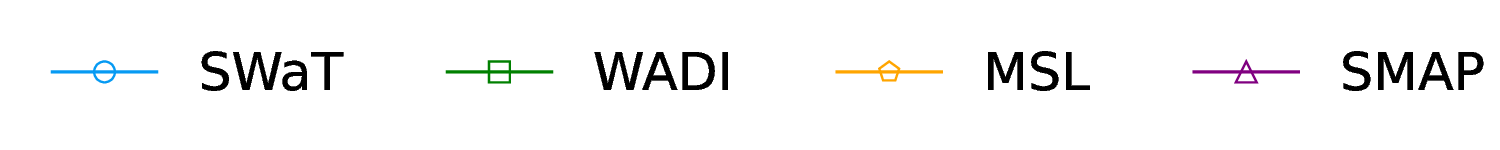}
    \vspace{-0.1cm}
  \end{subfigure}
  \newline
  \begin{subfigure}{0.24\linewidth}
  \centering
  \includegraphics[height=2.8cm]{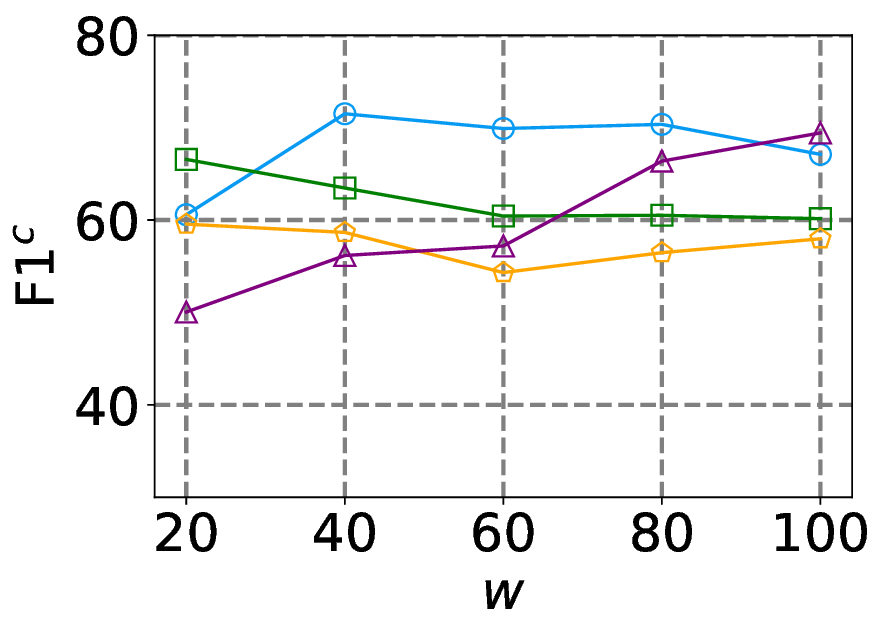}\label{fig: win_size_results}
  \caption{F1$^c$ vs. context window}
  \end{subfigure}
    \begin{subfigure}{0.24\linewidth}
  \centering
  \includegraphics[height=2.8cm]{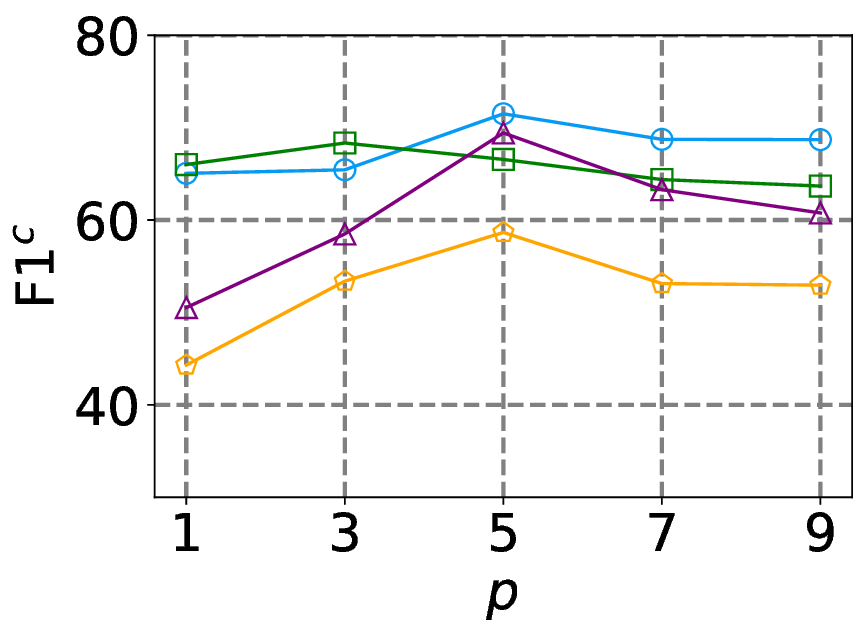}\label{fig: reconst_size_results}
  \caption{F1$^c$ vs. prediction window}
  \end{subfigure}
    \begin{subfigure}{0.24\linewidth}
  \centering
  \includegraphics[height=2.8cm]{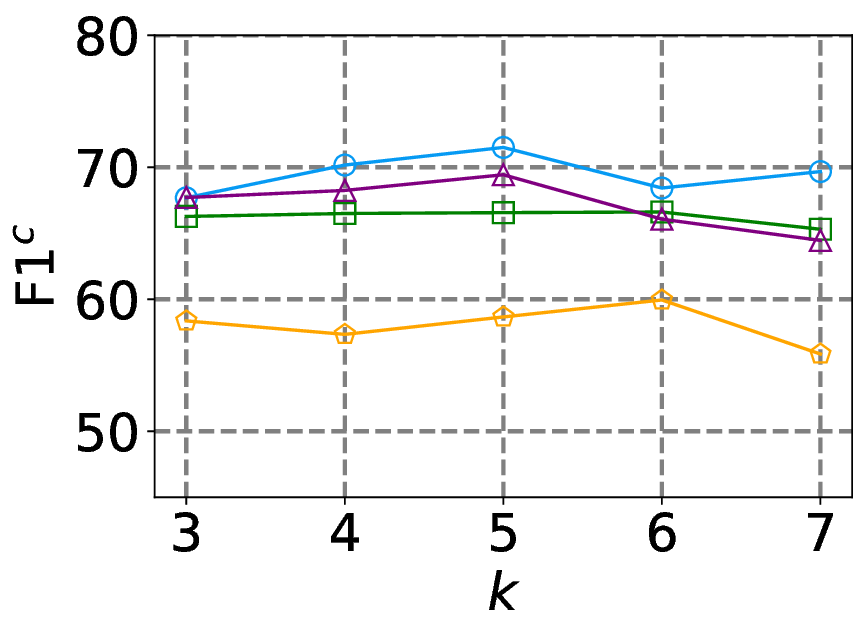}\label{fig: n_head_results}
  \caption{F1$^c$ vs. \# dynamic graphs}
  \end{subfigure}
  \begin{subfigure}{0.24\linewidth}
  \centering
  \includegraphics[height=2.8cm]{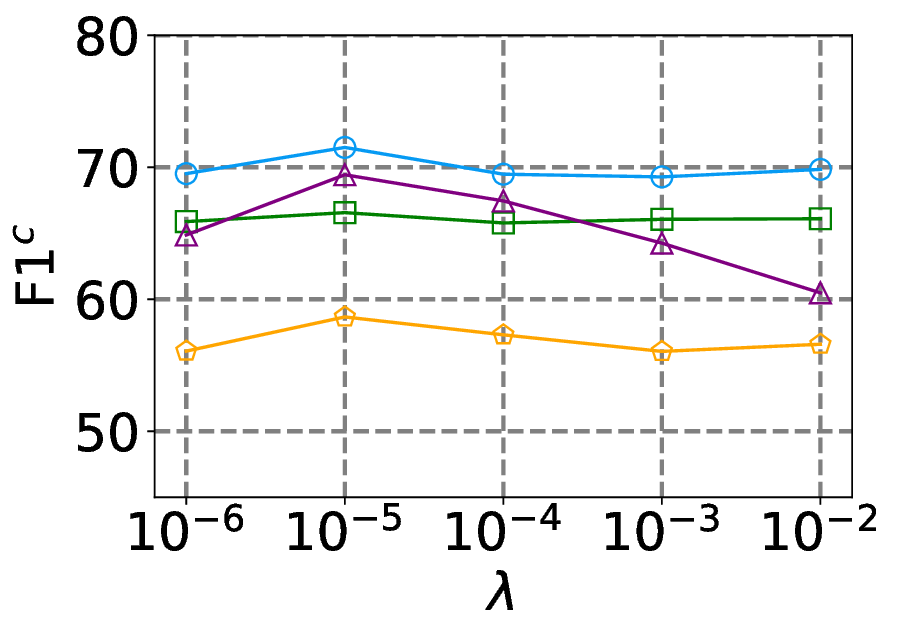}\label{fig: graph_dev_results}
  \caption{F1$^c$ vs. $\LL^{gc}$ weight}
  \end{subfigure}


\caption{Performance variation in terms of F1$^c$ under different hyper-parameters.}
\label{fig: sensitivity_results}
\end{figure*}


\subsection{Sensitivity Analysis}

Figure~\ref{fig: sensitivity_results} presents the impact of four parameters on the performance of PMGC. For the loss weights of the graph cohesion loss, performance is peaked at $10^{-5}$ and tends to decline slightly with larger weights. This outcome is understandable, as excessively high loss weights may overly prioritize graph diversification, detracting from the primary forecasting task, while too small weights can reduce the effectiveness of dynamic graph regularization. The number of graphs, $k$, shows optimal performance around 5, suggesting a moderate number of graphs can maintain a good anomaly detection performance. Furthermore, this relatively small number of graphs ensures high computational efficiency. 
For the context window size $w$, the performance varies across datasets. For instance, SMAP demonstrates improved performance with larger window sizes, while WADI exhibits a decline. These discrepancies may be attributed to the trade-off between information richness and sample correlation. Larger windows provide richer contextual information but may lead to overfitting due to increased sample-wise correlation, while smaller windows have the opposite effect Conversely. Overall, the choice of $w$ is dataset-specific and guided by validation forecasting performance.
Lastly, for the prediction window $p$, performance peaks around $p=5$, partly because incorporating more future steps better utilizes dynamic graphs, which in turn accurately reflects the current inter-series relationship among time series.

\section{Conclusion}

In this paper, we present a multi-graph based TSAD framework that integrates dynamic and static graphs to model both short-term and long-term inter-series relationships. A graph cohesion loss function is proposed, supported by theoretical analysis, which balances diversity and cohesiveness in graph representations. This ensures that dynamic graphs adhere to fundamental inter-series relationships while adapting to multifaceted short-term variations. Additionally, the ``prospective graphing'' technique substantially enhances predictive accuracy by leveraging concurrent information during the graph construction process. Empirical evaluations on five real-world industrial datasets demonstrated that PMGC significantly outperforms various existing baseline methods, showcasing its effectiveness in accurately detecting anomalies multivariate time series data. 


  \bibliographystyle{ACM-Reference-Format}
  \bibliography{mybibfile}


\begin{thebibliography}{48}


\ifx \showCODEN    \undefined \def \showCODEN     #1{\unskip}     \fi
\ifx \showDOI      \undefined \def \showDOI       #1{#1}\fi
\ifx \showISBNx    \undefined \def \showISBNx     #1{\unskip}     \fi
\ifx \showISBNxiii \undefined \def \showISBNxiii  #1{\unskip}     \fi
\ifx \showISSN     \undefined \def \showISSN      #1{\unskip}     \fi
\ifx \showLCCN     \undefined \def \showLCCN      #1{\unskip}     \fi
\ifx \shownote     \undefined \def \shownote      #1{#1}          \fi
\ifx \showarticletitle \undefined \def \showarticletitle #1{#1}   \fi
\ifx \showURL      \undefined \def \showURL       {\relax}        \fi
\providecommand\bibfield[2]{#2}
\providecommand\bibinfo[2]{#2}
\providecommand\natexlab[1]{#1}
\providecommand\showeprint[2][]{arXiv:#2}

\bibitem[Ahmed et~al\mbox{.}(2017)]%
        {ahmed2017wadi}
\bibfield{author}{\bibinfo{person}{Chuadhry~Mujeeb Ahmed}, \bibinfo{person}{Venkata~Reddy Palleti}, {and} \bibinfo{person}{Aditya~P Mathur}.} \bibinfo{year}{2017}\natexlab{}.
\newblock \showarticletitle{WADI: a water distribution testbed for research in the design of secure cyber physical systems}. In \bibinfo{booktitle}{\emph{International Workshop on Cyber-physical Systems for Smart Water Networks}}. \bibinfo{pages}{25--28}.
\newblock


\bibitem[Anandakrishnan et~al\mbox{.}(2018)]%
        {anandakrishnan2018anomaly}
\bibfield{author}{\bibinfo{person}{Archana Anandakrishnan}, \bibinfo{person}{Senthil Kumar}, \bibinfo{person}{Alexander Statnikov}, \bibinfo{person}{Tanveer Faruquie}, {and} \bibinfo{person}{Di Xu}.} \bibinfo{year}{2018}\natexlab{}.
\newblock \showarticletitle{Anomaly Detection in Finance: Editors’ Introduction}. In \bibinfo{booktitle}{\emph{Proceedings of the KDD 2017: Workshop on Anomaly Detection in Finance}}. PMLR, \bibinfo{pages}{1--7}.
\newblock


\bibitem[Bai et~al\mbox{.}(2018)]%
        {tcn}
\bibfield{author}{\bibinfo{person}{Shaojie Bai}, \bibinfo{person}{J~Zico Kolter}, {and} \bibinfo{person}{Vladlen Koltun}.} \bibinfo{year}{2018}\natexlab{}.
\newblock \showarticletitle{An empirical evaluation of generic convolutional and recurrent networks for sequence modeling}.
\newblock \bibinfo{journal}{\emph{arXiv preprint arXiv:1803.01271}} (\bibinfo{year}{2018}).
\newblock


\bibitem[Bashar and Nayak(2020)]%
        {tanogan}
\bibfield{author}{\bibinfo{person}{Md~Abul Bashar} {and} \bibinfo{person}{Richi Nayak}.} \bibinfo{year}{2020}\natexlab{}.
\newblock \showarticletitle{TAnoGAN: Time series anomaly detection with generative adversarial networks}. In \bibinfo{booktitle}{\emph{IEEE Symposium Series on Computational Intelligence}}. IEEE, \bibinfo{pages}{1778--1785}.
\newblock


\bibitem[Box et~al\mbox{.}(2015)]%
        {box2015time}
\bibfield{author}{\bibinfo{person}{George~EP Box}, \bibinfo{person}{Gwilym~M Jenkins}, \bibinfo{person}{Gregory~C Reinsel}, {and} \bibinfo{person}{Greta~M Ljung}.} \bibinfo{year}{2015}\natexlab{}.
\newblock \bibinfo{booktitle}{\emph{Time series analysis: forecasting and control}}.
\newblock \bibinfo{publisher}{John Wiley \& Sons}.
\newblock


\bibitem[Chen et~al\mbox{.}(2023b)]%
        {dygraphad}
\bibfield{author}{\bibinfo{person}{Jiazhen Chen}, \bibinfo{person}{Mingbin Feng}, {and} \bibinfo{person}{Tony~S Wirjanto}.} \bibinfo{year}{2023}\natexlab{b}.
\newblock \showarticletitle{Multivariate time series anomaly detection via dynamic graph forecasting}.
\newblock \bibinfo{journal}{\emph{arXiv preprint arXiv:2302.02051}} (\bibinfo{year}{2023}).
\newblock


\bibitem[Chen et~al\mbox{.}(2025a)]%
        {chen2025harnessing}
\bibfield{author}{\bibinfo{person}{Jiazhen Chen}, \bibinfo{person}{Mingbin Feng}, {and} \bibinfo{person}{Tony~S Wirjanto}.} \bibinfo{year}{2025}\natexlab{a}.
\newblock \showarticletitle{Harnessing contrastive learning and neural transformation for time series anomaly detection}. In \bibinfo{booktitle}{\emph{ICASSP 2025-2025 IEEE International Conference on Acoustics, Speech and Signal Processing (ICASSP)}}. IEEE, \bibinfo{pages}{1--5}.
\newblock


\bibitem[Chen et~al\mbox{.}(2025b)]%
        {chen2025semi}
\bibfield{author}{\bibinfo{person}{Jiazhen Chen}, \bibinfo{person}{Sichao Fu}, \bibinfo{person}{Zheng Ma}, \bibinfo{person}{Mingbin Feng}, \bibinfo{person}{Tony~S Wirjanto}, {and} \bibinfo{person}{Qinmu Peng}.} \bibinfo{year}{2025}\natexlab{b}.
\newblock \showarticletitle{Semi-supervised Anomaly Detection with Extremely Limited Labels in Dynamic Graphs}.
\newblock \bibinfo{journal}{\emph{arXiv preprint arXiv:2501.15035}} (\bibinfo{year}{2025}).
\newblock


\bibitem[Chen et~al\mbox{.}(2021b)]%
        {chen2021daemon}
\bibfield{author}{\bibinfo{person}{Xuanhao Chen}, \bibinfo{person}{Liwei Deng}, \bibinfo{person}{Feiteng Huang}, \bibinfo{person}{Chengwei Zhang}, \bibinfo{person}{Zongquan Zhang}, \bibinfo{person}{Yan Zhao}, {and} \bibinfo{person}{Kai Zheng}.} \bibinfo{year}{2021}\natexlab{b}.
\newblock \showarticletitle{Daemon: Unsupervised anomaly detection and interpretation for multivariate time series}. In \bibinfo{booktitle}{\emph{IEEE International Conference on Data Engineering}}. IEEE, \bibinfo{pages}{2225--2230}.
\newblock


\bibitem[Chen et~al\mbox{.}(2023a)]%
        {chen2023adversarial}
\bibfield{author}{\bibinfo{person}{Xuanhao Chen}, \bibinfo{person}{Liwei Deng}, \bibinfo{person}{Yan Zhao}, {and} \bibinfo{person}{Kai Zheng}.} \bibinfo{year}{2023}\natexlab{a}.
\newblock \showarticletitle{Adversarial autoencoder for unsupervised time series anomaly detection and interpretation}. In \bibinfo{booktitle}{\emph{ACM International Conference on Web Search and Data Mining}}. \bibinfo{pages}{267--275}.
\newblock


\bibitem[Chen et~al\mbox{.}(2021a)]%
        {gta}
\bibfield{author}{\bibinfo{person}{Zekai Chen}, \bibinfo{person}{Dingshuo Chen}, \bibinfo{person}{Xiao Zhang}, \bibinfo{person}{Zixuan Yuan}, {and} \bibinfo{person}{Xiuzhen Cheng}.} \bibinfo{year}{2021}\natexlab{a}.
\newblock \showarticletitle{Learning graph structures with transformer for multivariate time-series anomaly detection in IoT}.
\newblock \bibinfo{journal}{\emph{IEEE Internet of Things Journal}} \bibinfo{volume}{9}, \bibinfo{number}{12} (\bibinfo{year}{2021}), \bibinfo{pages}{9179--9189}.
\newblock


\bibitem[Cook et~al\mbox{.}(2019)]%
        {cook2019anomaly}
\bibfield{author}{\bibinfo{person}{Andrew~A Cook}, \bibinfo{person}{G{\"o}ksel M{\i}s{\i}rl{\i}}, {and} \bibinfo{person}{Zhong Fan}.} \bibinfo{year}{2019}\natexlab{}.
\newblock \showarticletitle{Anomaly detection for IoT time-series data: A survey}.
\newblock \bibinfo{journal}{\emph{IEEE Internet of Things Journal}} \bibinfo{volume}{7}, \bibinfo{number}{7} (\bibinfo{year}{2019}), \bibinfo{pages}{6481--6494}.
\newblock


\bibitem[Deng and Hooi(2021)]%
        {gdn}
\bibfield{author}{\bibinfo{person}{Ailin Deng} {and} \bibinfo{person}{Bryan Hooi}.} \bibinfo{year}{2021}\natexlab{}.
\newblock \showarticletitle{Graph neural network-based anomaly detection in multivariate time series}. In \bibinfo{booktitle}{\emph{AAAI Conference on Artificial Intelligence}}, Vol.~\bibinfo{volume}{35}. \bibinfo{pages}{4027--4035}.
\newblock


\bibitem[Fang et~al\mbox{.}(2024)]%
        {fang2024temporal}
\bibfield{author}{\bibinfo{person}{Yuchen Fang}, \bibinfo{person}{Jiandong Xie}, \bibinfo{person}{Yan Zhao}, \bibinfo{person}{Lu Chen}, \bibinfo{person}{Yunjun Gao}, {and} \bibinfo{person}{Kai Zheng}.} \bibinfo{year}{2024}\natexlab{}.
\newblock \showarticletitle{Temporal-frequency masked autoencoders for time series anomaly detection}. In \bibinfo{booktitle}{\emph{International Conference on Data Engineering}}. IEEE, \bibinfo{pages}{1228--1241}.
\newblock


\bibitem[Garg et~al\mbox{.}(2021)]%
        {garg2021evaluation}
\bibfield{author}{\bibinfo{person}{Astha Garg}, \bibinfo{person}{Wenyu Zhang}, \bibinfo{person}{Jules Samaran}, \bibinfo{person}{Ramasamy Savitha}, {and} \bibinfo{person}{Chuan-Sheng Foo}.} \bibinfo{year}{2021}\natexlab{}.
\newblock \showarticletitle{An evaluation of anomaly detection and diagnosis in multivariate time series}.
\newblock \bibinfo{journal}{\emph{IEEE Transactions on Neural Networks and Learning Systems}} \bibinfo{volume}{33}, \bibinfo{number}{6} (\bibinfo{year}{2021}), \bibinfo{pages}{2508--2517}.
\newblock


\bibitem[Han and Woo(2022)]%
        {fusagnet}
\bibfield{author}{\bibinfo{person}{Siho Han} {and} \bibinfo{person}{Simon~S Woo}.} \bibinfo{year}{2022}\natexlab{}.
\newblock \showarticletitle{Learning sparse latent graph representations for anomaly detection in multivariate time series}. In \bibinfo{booktitle}{\emph{ACM SIGKDD Conference on Knowledge Discovery and Data Mining}}. \bibinfo{pages}{2977--2986}.
\newblock


\bibitem[He and Zhao(2019)]%
        {tcnanomaly}
\bibfield{author}{\bibinfo{person}{Yangdong He} {and} \bibinfo{person}{Jiabao Zhao}.} \bibinfo{year}{2019}\natexlab{}.
\newblock \showarticletitle{Temporal convolutional networks for anomaly detection in time series}. In \bibinfo{booktitle}{\emph{Journal of Physics: Conference Series}}, Vol.~\bibinfo{volume}{1213}. IOP Publishing, \bibinfo{pages}{042050}.
\newblock


\bibitem[Hundman et~al\mbox{.}(2018)]%
        {lstm-ndt}
\bibfield{author}{\bibinfo{person}{Kyle Hundman}, \bibinfo{person}{Valentino Constantinou}, \bibinfo{person}{Christopher Laporte}, \bibinfo{person}{Ian Colwell}, {and} \bibinfo{person}{Tom Soderstrom}.} \bibinfo{year}{2018}\natexlab{}.
\newblock \showarticletitle{Detecting spacecraft anomalies using lstms and nonparametric dynamic thresholding}. In \bibinfo{booktitle}{\emph{ACM SIGKDD International Conference on Knowledge Discovery and Data Mining}}. \bibinfo{pages}{387--395}.
\newblock


\bibitem[Jie et~al\mbox{.}(2024)]%
        {jie2024disentangled}
\bibfield{author}{\bibinfo{person}{Xin Jie}, \bibinfo{person}{Xixi Zhou}, \bibinfo{person}{Chanfei Su}, \bibinfo{person}{Zijun Zhou}, \bibinfo{person}{Yuqing Yuan}, \bibinfo{person}{Jiajun Bu}, {and} \bibinfo{person}{Haishuai Wang}.} \bibinfo{year}{2024}\natexlab{}.
\newblock \showarticletitle{Disentangled anomaly detection for multivariate time series}. In \bibinfo{booktitle}{\emph{Companion Proceedings of the ACM on Web Conference}}. \bibinfo{pages}{931--934}.
\newblock


\bibitem[Kim et~al\mbox{.}(2022)]%
        {kim2022towards}
\bibfield{author}{\bibinfo{person}{Siwon Kim}, \bibinfo{person}{Kukjin Choi}, \bibinfo{person}{Hyun-Soo Choi}, \bibinfo{person}{Byunghan Lee}, {and} \bibinfo{person}{Sungroh Yoon}.} \bibinfo{year}{2022}\natexlab{}.
\newblock In \bibinfo{booktitle}{\emph{AAAI Conference on Artificial Intelligence}}, Vol.~\bibinfo{volume}{36}. \bibinfo{pages}{7194--7201}.
\newblock


\bibitem[Kingma and Ba(2014)]%
        {kingma2014adam}
\bibfield{author}{\bibinfo{person}{Diederik~P Kingma} {and} \bibinfo{person}{Jimmy Ba}.} \bibinfo{year}{2014}\natexlab{}.
\newblock \showarticletitle{Adam: A method for stochastic optimization}.
\newblock \bibinfo{journal}{\emph{arXiv preprint arXiv:1412.6980}} (\bibinfo{year}{2014}).
\newblock


\bibitem[Lee et~al\mbox{.}(2023)]%
        {lee2023duogat}
\bibfield{author}{\bibinfo{person}{Jongsoo Lee}, \bibinfo{person}{Byeongtae Park}, {and} \bibinfo{person}{Dong-Kyu Chae}.} \bibinfo{year}{2023}\natexlab{}.
\newblock \showarticletitle{DuoGAT: Dual time-oriented graph attention networks for accurate, efficient and explainable anomaly detection on time-series}. In \bibinfo{booktitle}{\emph{ACM International Conference on Information and Knowledge Management}}. \bibinfo{pages}{1188--1197}.
\newblock


\bibitem[Li et~al\mbox{.}(2019)]%
        {li2019mad}
\bibfield{author}{\bibinfo{person}{Dan Li}, \bibinfo{person}{Dacheng Chen}, \bibinfo{person}{Baihong Jin}, \bibinfo{person}{Lei Shi}, \bibinfo{person}{Jonathan Goh}, {and} \bibinfo{person}{See-Kiong Ng}.} \bibinfo{year}{2019}\natexlab{}.
\newblock \showarticletitle{MAD-GAN: Multivariate anomaly detection for time series data with generative adversarial networks}. In \bibinfo{booktitle}{\emph{International Conference on Artificial Neural Networks}}. Springer, \bibinfo{pages}{703--716}.
\newblock


\bibitem[Li et~al\mbox{.}(2021)]%
        {li2021fluxev}
\bibfield{author}{\bibinfo{person}{Jia Li}, \bibinfo{person}{Shimin Di}, \bibinfo{person}{Yanyan Shen}, {and} \bibinfo{person}{Lei Chen}.} \bibinfo{year}{2021}\natexlab{}.
\newblock \showarticletitle{FluxEV: a fast and effective unsupervised framework for time-series anomaly detection}. In \bibinfo{booktitle}{\emph{ACM International Conference on Web Search and Data Mining}}. \bibinfo{pages}{824--832}.
\newblock


\bibitem[Li et~al\mbox{.}(2023)]%
        {li2023prototype}
\bibfield{author}{\bibinfo{person}{Yuxin Li}, \bibinfo{person}{Wenchao Chen}, \bibinfo{person}{Bo Chen}, \bibinfo{person}{Dongsheng Wang}, \bibinfo{person}{Long Tian}, {and} \bibinfo{person}{Mingyuan Zhou}.} \bibinfo{year}{2023}\natexlab{}.
\newblock \showarticletitle{Prototype-oriented unsupervised anomaly detection for multivariate time series}. In \bibinfo{booktitle}{\emph{International Conference on Machine Learning}}. PMLR, \bibinfo{pages}{19407--19424}.
\newblock


\bibitem[Malhotra et~al\mbox{.}(2016)]%
        {lstmed}
\bibfield{author}{\bibinfo{person}{Pankaj Malhotra}, \bibinfo{person}{Anusha Ramakrishnan}, \bibinfo{person}{Gaurangi Anand}, \bibinfo{person}{Lovekesh Vig}, \bibinfo{person}{Puneet Agarwal}, {and} \bibinfo{person}{Gautam Shroff}.} \bibinfo{year}{2016}\natexlab{}.
\newblock \showarticletitle{LSTM-based encoder-decoder for multi-sensor anomaly detection}.
\newblock \bibinfo{journal}{\emph{arXiv preprint arXiv:1607.00148}} (\bibinfo{year}{2016}).
\newblock


\bibitem[Mathur and Tippenhauer(2016)]%
        {mathur2016swat}
\bibfield{author}{\bibinfo{person}{Aditya~P Mathur} {and} \bibinfo{person}{Nils~Ole Tippenhauer}.} \bibinfo{year}{2016}\natexlab{}.
\newblock \showarticletitle{SWaT: A water treatment testbed for research and training on ICS security}. In \bibinfo{booktitle}{\emph{International Workshop on Cyber-physical Systems for Smart Water Networks}}. IEEE, \bibinfo{pages}{31--36}.
\newblock


\bibitem[Munir et~al\mbox{.}(2018)]%
        {munir2018deepant}
\bibfield{author}{\bibinfo{person}{Mohsin Munir}, \bibinfo{person}{Shoaib~Ahmed Siddiqui}, \bibinfo{person}{Andreas Dengel}, {and} \bibinfo{person}{Sheraz Ahmed}.} \bibinfo{year}{2018}\natexlab{}.
\newblock \showarticletitle{DeepAnT: A deep learning approach for unsupervised anomaly detection in time series}.
\newblock \bibinfo{journal}{\emph{IEEE Access}}  \bibinfo{volume}{7} (\bibinfo{year}{2018}), \bibinfo{pages}{1991--2005}.
\newblock


\bibitem[Paparrizos et~al\mbox{.}(2022)]%
        {paparrizos2022volume}
\bibfield{author}{\bibinfo{person}{John Paparrizos}, \bibinfo{person}{Paul Boniol}, \bibinfo{person}{Themis Palpanas}, \bibinfo{person}{Ruey~S Tsay}, \bibinfo{person}{Aaron Elmore}, {and} \bibinfo{person}{Michael~J Franklin}.} \bibinfo{year}{2022}\natexlab{}.
\newblock \showarticletitle{Volume under the surface: a new accuracy evaluation measure for time-series anomaly detection}. In \bibinfo{booktitle}{\emph{VLDB Endowment}}.
\newblock


\bibitem[Qi et~al\mbox{.}(2022)]%
        {mad-sgcn}
\bibfield{author}{\bibinfo{person}{Panpan Qi}, \bibinfo{person}{Dan Li}, {and} \bibinfo{person}{See-Kiong Ng}.} \bibinfo{year}{2022}\natexlab{}.
\newblock \showarticletitle{MAD-SGCN: Multivariate anomaly detection with self-learning graph convolutional networks}. In \bibinfo{booktitle}{\emph{International Conference on Data Engineering}}. IEEE, \bibinfo{pages}{1232--1244}.
\newblock


\bibitem[Ren et~al\mbox{.}(2019)]%
        {ren2019time}
\bibfield{author}{\bibinfo{person}{Hansheng Ren}, \bibinfo{person}{Bixiong Xu}, \bibinfo{person}{Yujing Wang}, \bibinfo{person}{Chao Yi}, \bibinfo{person}{Congrui Huang}, \bibinfo{person}{Xiaoyu Kou}, \bibinfo{person}{Tony Xing}, \bibinfo{person}{Mao Yang}, \bibinfo{person}{Jie Tong}, {and} \bibinfo{person}{Qi Zhang}.} \bibinfo{year}{2019}\natexlab{}.
\newblock \showarticletitle{Time-series anomaly detection service at microsoft}. In \bibinfo{booktitle}{\emph{ACM SIGKDD International Conference on Knowledge Discovery and Data Mining}}. \bibinfo{pages}{3009--3017}.
\newblock


\bibitem[Sch{\"o}lkopf et~al\mbox{.}(2001)]%
        {ocsvm}
\bibfield{author}{\bibinfo{person}{Bernhard Sch{\"o}lkopf}, \bibinfo{person}{John~C Platt}, \bibinfo{person}{John Shawe-Taylor}, \bibinfo{person}{Alex~J Smola}, {and} \bibinfo{person}{Robert~C Williamson}.} \bibinfo{year}{2001}\natexlab{}.
\newblock \showarticletitle{Estimating the support of a high-dimensional distribution}.
\newblock \bibinfo{journal}{\emph{Neural Computation}} \bibinfo{volume}{13}, \bibinfo{number}{7} (\bibinfo{year}{2001}), \bibinfo{pages}{1443--1471}.
\newblock


\bibitem[Shin et~al\mbox{.}(2021)]%
        {hai}
\bibfield{author}{\bibinfo{person}{Hyeok-Ki Shin}, \bibinfo{person}{Woomyo Lee}, \bibinfo{person}{Jeong-Han Yun}, {and} \bibinfo{person}{Byung-Gi Min}.} \bibinfo{year}{2021}\natexlab{}.
\newblock \showarticletitle{Two ICS security datasets and anomaly detection contest on the HIL-based augmented ICS testbed}. In \bibinfo{booktitle}{\emph{Cyber Security Experimentation and Test Workshop}} (Virtual, CA, USA) \emph{(\bibinfo{series}{CSET '21})}. \bibinfo{publisher}{Association for Computing Machinery}, \bibinfo{address}{New York, NY, USA}, \bibinfo{pages}{36–40}.
\newblock
\showISBNx{9781450390651}
\urldef\tempurl%
\url{https://doi.org/10.1145/3474718.3474719}
\showDOI{\tempurl}


\bibitem[Shyu et~al\mbox{.}(2003)]%
        {shyu2003novel}
\bibfield{author}{\bibinfo{person}{Mei-Ling Shyu}, \bibinfo{person}{Shu-Ching Chen}, \bibinfo{person}{Kanoksri Sarinnapakorn}, {and} \bibinfo{person}{LiWu Chang}.} \bibinfo{year}{2003}\natexlab{}.
\newblock \showarticletitle{A novel anomaly detection scheme based on principal component classifier}. In \bibinfo{booktitle}{\emph{IEEE Foundations and New Directions of Data Mining Workshop}}. IEEE Press, \bibinfo{pages}{172--179}.
\newblock


\bibitem[Su et~al\mbox{.}(2019)]%
        {omnianomaly}
\bibfield{author}{\bibinfo{person}{Ya Su}, \bibinfo{person}{Youjian Zhao}, \bibinfo{person}{Chenhao Niu}, \bibinfo{person}{Rong Liu}, \bibinfo{person}{Wei Sun}, {and} \bibinfo{person}{Dan Pei}.} \bibinfo{year}{2019}\natexlab{}.
\newblock \showarticletitle{Robust anomaly detection for multivariate time series through stochastic recurrent neural network}. In \bibinfo{booktitle}{\emph{ACM SIGKDD International Conference on Knowledge Discovery and Data Mining}}. \bibinfo{pages}{2828--2837}.
\newblock


\bibitem[Tian et~al\mbox{.}(2023)]%
        {tian2023anomaly}
\bibfield{author}{\bibinfo{person}{Zhiwen Tian}, \bibinfo{person}{Ming Zhuo}, \bibinfo{person}{Leyuan Liu}, \bibinfo{person}{Junyi Chen}, {and} \bibinfo{person}{Shijie Zhou}.} \bibinfo{year}{2023}\natexlab{}.
\newblock \showarticletitle{Anomaly detection using spatial and temporal information in multivariate time series}.
\newblock \bibinfo{journal}{\emph{Scientific Reports}} \bibinfo{volume}{13}, \bibinfo{number}{1} (\bibinfo{year}{2023}), \bibinfo{pages}{4400}.
\newblock


\bibitem[Tuli et~al\mbox{.}(2022)]%
        {tuli2022tranad}
\bibfield{author}{\bibinfo{person}{Shreshth Tuli}, \bibinfo{person}{Giuliano Casale}, {and} \bibinfo{person}{Nicholas~R Jennings}.} \bibinfo{year}{2022}\natexlab{}.
\newblock \showarticletitle{TranAD: deep transformer networks for anomaly detection in multivariate time series data}. In \bibinfo{booktitle}{\emph{VLDB}}.
\newblock


\bibitem[Vaswani et~al\mbox{.}(2017)]%
        {transformer}
\bibfield{author}{\bibinfo{person}{Ashish Vaswani}, \bibinfo{person}{Noam Shazeer}, \bibinfo{person}{Niki Parmar}, \bibinfo{person}{Jakob Uszkoreit}, \bibinfo{person}{Llion Jones}, \bibinfo{person}{Aidan~N Gomez}, \bibinfo{person}{{\L}ukasz Kaiser}, {and} \bibinfo{person}{Illia Polosukhin}.} \bibinfo{year}{2017}\natexlab{}.
\newblock \showarticletitle{Attention is all you need}.
\newblock \bibinfo{journal}{\emph{Advances in Neural Information Processing Systems}}  \bibinfo{volume}{30} (\bibinfo{year}{2017}).
\newblock


\bibitem[Wang et~al\mbox{.}(2022)]%
        {wang2022multiscale}
\bibfield{author}{\bibinfo{person}{Jing Wang}, \bibinfo{person}{Shikuan Shao}, \bibinfo{person}{Yunfei Bai}, \bibinfo{person}{Jiaoxue Deng}, {and} \bibinfo{person}{Youfang Lin}.} \bibinfo{year}{2022}\natexlab{}.
\newblock \showarticletitle{Multiscale wavelet graph autoEncoder for multivariate time-series anomaly detection}.
\newblock \bibinfo{journal}{\emph{IEEE Transactions on Instrumentation and Measurement}}  \bibinfo{volume}{72} (\bibinfo{year}{2022}), \bibinfo{pages}{1--11}.
\newblock


\bibitem[Wang et~al\mbox{.}(2023)]%
        {wang2023deep}
\bibfield{author}{\bibinfo{person}{Rui Wang}, \bibinfo{person}{Chongwei Liu}, \bibinfo{person}{Xudong Mou}, \bibinfo{person}{Kai Gao}, \bibinfo{person}{Xiaohui Guo}, \bibinfo{person}{Pin Liu}, \bibinfo{person}{Tianyu Wo}, {and} \bibinfo{person}{Xudong Liu}.} \bibinfo{year}{2023}\natexlab{}.
\newblock \showarticletitle{Deep contrastive one-class time series anomaly detection}. In \bibinfo{booktitle}{\emph{SIAM International Conference on Data Mining}}. SIAM, \bibinfo{pages}{694--702}.
\newblock


\bibitem[Wu et~al\mbox{.}(2020)]%
        {wu2020connecting}
\bibfield{author}{\bibinfo{person}{Zonghan Wu}, \bibinfo{person}{Shirui Pan}, \bibinfo{person}{Guodong Long}, \bibinfo{person}{Jing Jiang}, \bibinfo{person}{Xiaojun Chang}, {and} \bibinfo{person}{Chengqi Zhang}.} \bibinfo{year}{2020}\natexlab{}.
\newblock \showarticletitle{Connecting the dots: Multivariate time series forecasting with graph neural networks}. In \bibinfo{booktitle}{\emph{ACM SIGKDD International Conference on Knowledge Discovery and Data Mining}}. \bibinfo{pages}{753--763}.
\newblock


\bibitem[Xu et~al\mbox{.}(2022)]%
        {anomalyTransformer}
\bibfield{author}{\bibinfo{person}{Jiehui Xu}, \bibinfo{person}{Haixu Wu}, \bibinfo{person}{Jianmin Wang}, {and} \bibinfo{person}{Mingsheng Long}.} \bibinfo{year}{2022}\natexlab{}.
\newblock \showarticletitle{Anomaly Transformer: Time series anomaly detection with association discrepancy}. In \bibinfo{booktitle}{\emph{International Conference on Learning Representations}}.
\newblock


\bibitem[Yang et~al\mbox{.}(2023)]%
        {yang2023dcdetector}
\bibfield{author}{\bibinfo{person}{Yiyuan Yang}, \bibinfo{person}{Chaoli Zhang}, \bibinfo{person}{Tian Zhou}, \bibinfo{person}{Qingsong Wen}, {and} \bibinfo{person}{Liang Sun}.} \bibinfo{year}{2023}\natexlab{}.
\newblock \showarticletitle{DCdetector: Dual attention contrastive representation learning for time series anomaly detection}. In \bibinfo{booktitle}{\emph{ACM SIGKDD International Conference on Knowledge Discovery and Data Mining}}. \bibinfo{pages}{3033–3045}.
\newblock


\bibitem[Zeng et~al\mbox{.}(2023)]%
        {DLinear}
\bibfield{author}{\bibinfo{person}{Ailing Zeng}, \bibinfo{person}{Muxi Chen}, \bibinfo{person}{Lei Zhang}, {and} \bibinfo{person}{Qiang Xu}.} \bibinfo{year}{2023}\natexlab{}.
\newblock \showarticletitle{Are transformers effective for time series forecasting?}. In \bibinfo{booktitle}{\emph{AAAI Conference on Artificial Intelligence}}, Vol.~\bibinfo{volume}{37}. \bibinfo{pages}{11121--11128}.
\newblock


\bibitem[Zhang et~al\mbox{.}(2019)]%
        {mscred}
\bibfield{author}{\bibinfo{person}{Chuxu Zhang}, \bibinfo{person}{Dongjin Song}, \bibinfo{person}{Yuncong Chen}, \bibinfo{person}{Xinyang Feng}, \bibinfo{person}{Cristian Lumezanu}, \bibinfo{person}{Wei Cheng}, \bibinfo{person}{Jingchao Ni}, \bibinfo{person}{Bo Zong}, \bibinfo{person}{Haifeng Chen}, {and} \bibinfo{person}{Nitesh~V Chawla}.} \bibinfo{year}{2019}\natexlab{}.
\newblock \showarticletitle{A deep neural network for unsupervised anomaly detection and diagnosis in multivariate time series data}. In \bibinfo{booktitle}{\emph{AAAI Conference on Artificial Intelligence}}, Vol.~\bibinfo{volume}{33}. \bibinfo{pages}{1409--1416}.
\newblock


\bibitem[Zhang et~al\mbox{.}(2022)]%
        {zhang2022grelen}
\bibfield{author}{\bibinfo{person}{Weiqi Zhang}, \bibinfo{person}{Chen Zhang}, {and} \bibinfo{person}{Fugee Tsung}.} \bibinfo{year}{2022}\natexlab{}.
\newblock \showarticletitle{Grelen: Multivariate time series anomaly detection from the perspective of graph relational learning}. In \bibinfo{booktitle}{\emph{International Joint Conference on Artificial Intelligence}}. \bibinfo{pages}{2390--2397}.
\newblock


\bibitem[Zhao et~al\mbox{.}(2020)]%
        {zhao2020multivariate}
\bibfield{author}{\bibinfo{person}{Hang Zhao}, \bibinfo{person}{Yujing Wang}, \bibinfo{person}{Juanyong Duan}, \bibinfo{person}{Congrui Huang}, \bibinfo{person}{Defu Cao}, \bibinfo{person}{Yunhai Tong}, \bibinfo{person}{Bixiong Xu}, \bibinfo{person}{Jing Bai}, \bibinfo{person}{Jie Tong}, {and} \bibinfo{person}{Qi Zhang}.} \bibinfo{year}{2020}\natexlab{}.
\newblock \showarticletitle{Multivariate time-series anomaly detection via graph attention network}. In \bibinfo{booktitle}{\emph{IEEE International Conference on Data Mining}}. IEEE, \bibinfo{pages}{841--850}.
\newblock


\bibitem[Zheng et~al\mbox{.}(2023)]%
        {zheng2023correlation}
\bibfield{author}{\bibinfo{person}{Yu Zheng}, \bibinfo{person}{Huan~Yee Koh}, \bibinfo{person}{Ming Jin}, \bibinfo{person}{Lianhua Chi}, \bibinfo{person}{Khoa~T Phan}, \bibinfo{person}{Shirui Pan}, \bibinfo{person}{Yi-Ping~Phoebe Chen}, {and} \bibinfo{person}{Wei Xiang}.} \bibinfo{year}{2023}\natexlab{}.
\newblock \showarticletitle{Correlation-aware spatial-temporal graph learning for multivariate time-series anomaly detection}.
\newblock \bibinfo{journal}{\emph{IEEE Transactions on Neural Networks and Learning Systems}} (\bibinfo{year}{2023}).
\newblock


\end{thebibliography}
\end{document}